%% file: eccv2020submissionCR.tex
\documentclass[runningheads]{llncs}

\usepackage{subfig}
\usepackage[labelsep=period]{caption}

\usepackage{graphicx}

\usepackage{tikz}
\usepackage{comment} 
\usepackage{amsmath,amssymb} 
\usepackage{color}

\usepackage{booktabs}
\usepackage{tabu}
\usepackage{multirow}
\usepackage{enumitem}
\usepackage{algorithm,algorithmicx,algpseudocode}

\usepackage{xcolor,colortbl}
\definecolor{LightGray}{rgb}{0.8,0.8,0.8}

\newcommand{\eg}{\textit{e.g.}}
\newcommand{\wrt}{\textit{w.r.t.} }
\newcommand{\norm}[1]{\left\lVert#1\right\rVert}
\algrenewcommand{\algorithmiccomment}[1]{\hfill$\blacktriangleright$ #1}

\begin{document}
\pagestyle{headings}
\mainmatter
\def\ECCVSubNumber{100}  

\title{BroadFace: Looking at Tens of Thousands of People at Once for Face Recognition} 

\titlerunning{BroadFace for Face Recognition}
%

\makeatletter
\newcommand{\printfnsymbol}[1]{%
  \textsuperscript{\@fnsymbol{#1}}%
}
\makeatother

\author{
Yonghyun Kim \thanks{Equal contribution} \inst{1}\orcidID{0000-0003-0038-7850} \and
Wonpyo Park \printfnsymbol{1} \inst{2}\orcidID{0000-0003-0675-6362} \and
Jongju Shin\inst{1}\orcidID{0000-0002-7359-5258}}
\authorrunning{Y. Kim, W. Park and J. Shin}
%
\institute{Kakao Enterprise, Seongnam, Korea 
\\
\email{\{aiden.d, isaac.giant\}@kakaoenterprise.com} \and
Kakao Corp., Seongnam, Korea
\\
\email{tony.nn@kakaocorp.com}\\
}
\maketitle

\input{section/abstract}

\input{section/introduction}

\input{section/relatedworks}

\input{section/method}

\input{section/experiment}

\input{section/conclusion}

\section*{Acknowledgement}
We would like to thank AI R\&D team of Kakao Enterprise for the helpful discussion.
In particular, we would like to thank Yunmo Park who designed the visual materials.

\clearpage
%
%
\bibliographystyle{splncs04}
\bibliography{egbib}
\end{document}

%% file: section/abstract.tex
\begin{abstract}
The datasets of face recognition contain an enormous number of identities and instances.
However, conventional methods have difficulty in reflecting the entire distribution of the datasets because a mini-batch of small size contains only a small portion of all identities.
To overcome this difficulty, we propose a novel method called BroadFace, which is a learning process to consider a massive set of identities, comprehensively.
In BroadFace, a linear classifier learns optimal decision boundaries among identities from a large number of embedding vectors accumulated over past iterations.
By referring more instances at once, the optimality of the classifier is naturally increased on the entire datasets.
Thus, the encoder is also globally optimized by referring the weight matrix of the classifier.
Moreover, we propose a novel compensation method to increase the number of referenced instances in the training stage.
BroadFace can be easily applied on many existing methods to accelerate a learning process and obtain a significant improvement in accuracy without extra computational burden at inference stage.
We perform extensive ablation studies and experiments on various datasets to show the effectiveness of BroadFace, and also empirically prove the validity of our compensation method.
BroadFace achieves the \textit{state-of-the-art} results with significant improvements on nine datasets in 1:1 face verification and 1:N face identification tasks, and is also effective in image retrieval.
\keywords{face recognition, large mini-batch learning, image retrieval}
\end{abstract}

%% file: section/introduction.tex
\section{Introduction}

Face recognition is a key technique for many applications of biometric authentication such as electronic payment, lock screen of smartphones, and video surveillance.
The main tasks of face recognition are categorized into face verification and face identification.
In face verification, a pair of faces are compared to verify whether their identities are the same or different.
In face identification, the identity of a given face is determined by comparing it to a pre-registered gallery of identities.
Many researches \cite{10.1007/978-3-540-24670-1_36,6619233,joint_bayesian,5459250,CSML,fisherface,eigenface,WHT:ECCVW08:DBMW,QiYin:2011:AMF:2191740.2192084} on face recognition have been conducted for decades. The recent adoption \cite{ArcFace,marginal_loss,SphereFace,FaceNet,deepface,NormFace,CosFace,centerloss} of Convolutional Neural Networks (CNNs) has dramatically increased recognition accuracy.
However, many difficulties of face recognition still remain to be solved.

\begin{figure}[t]
	\begin{center}
		\subfloat[Typical Mini-batch Learning]{
			\label{fig:teaser_a}
			\includegraphics[height=4.5cm]{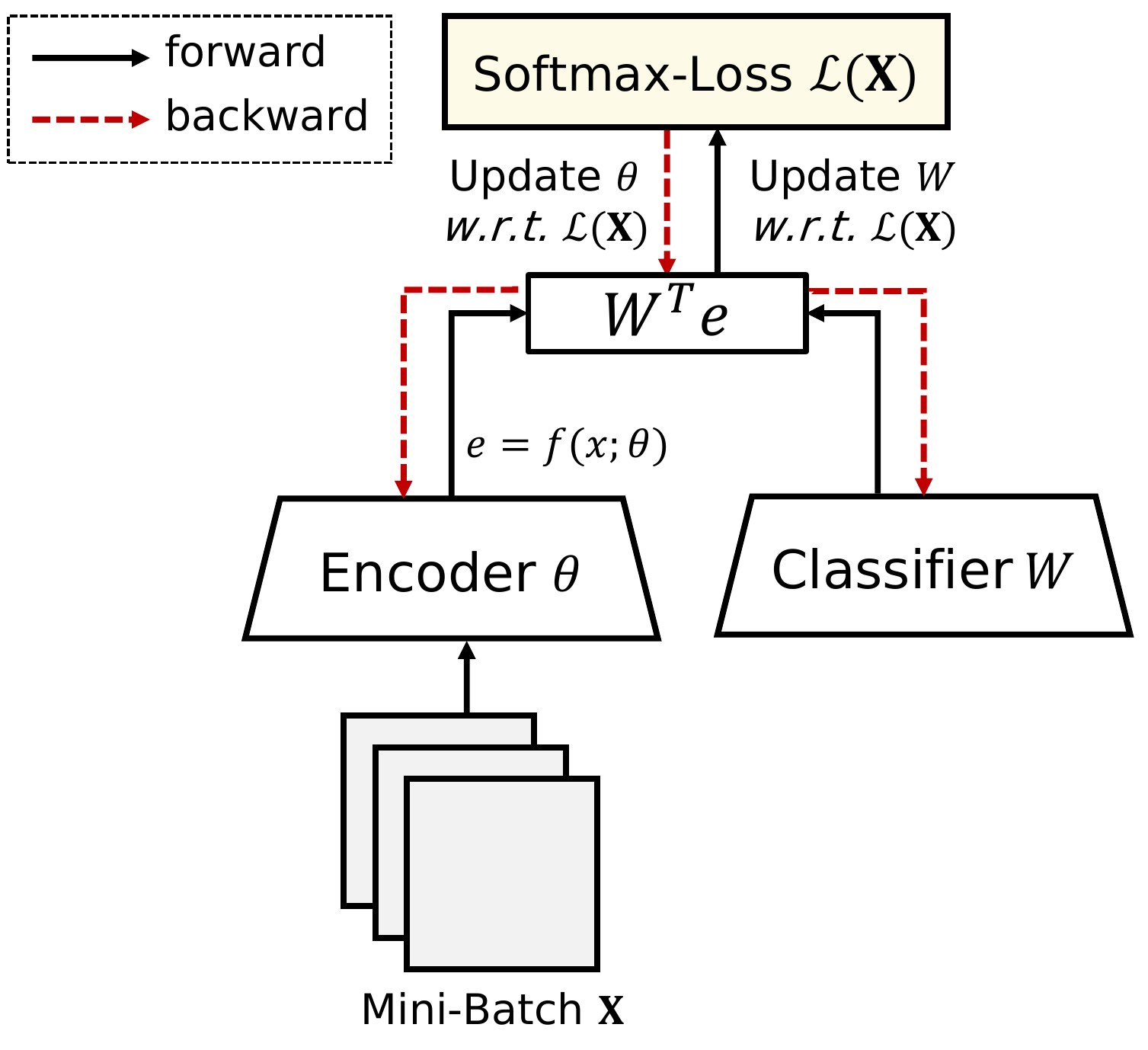}
		}
		\subfloat[The Proposed Learning]{
			\label{fig:teaser_b}
			\includegraphics[height=4.5cm]{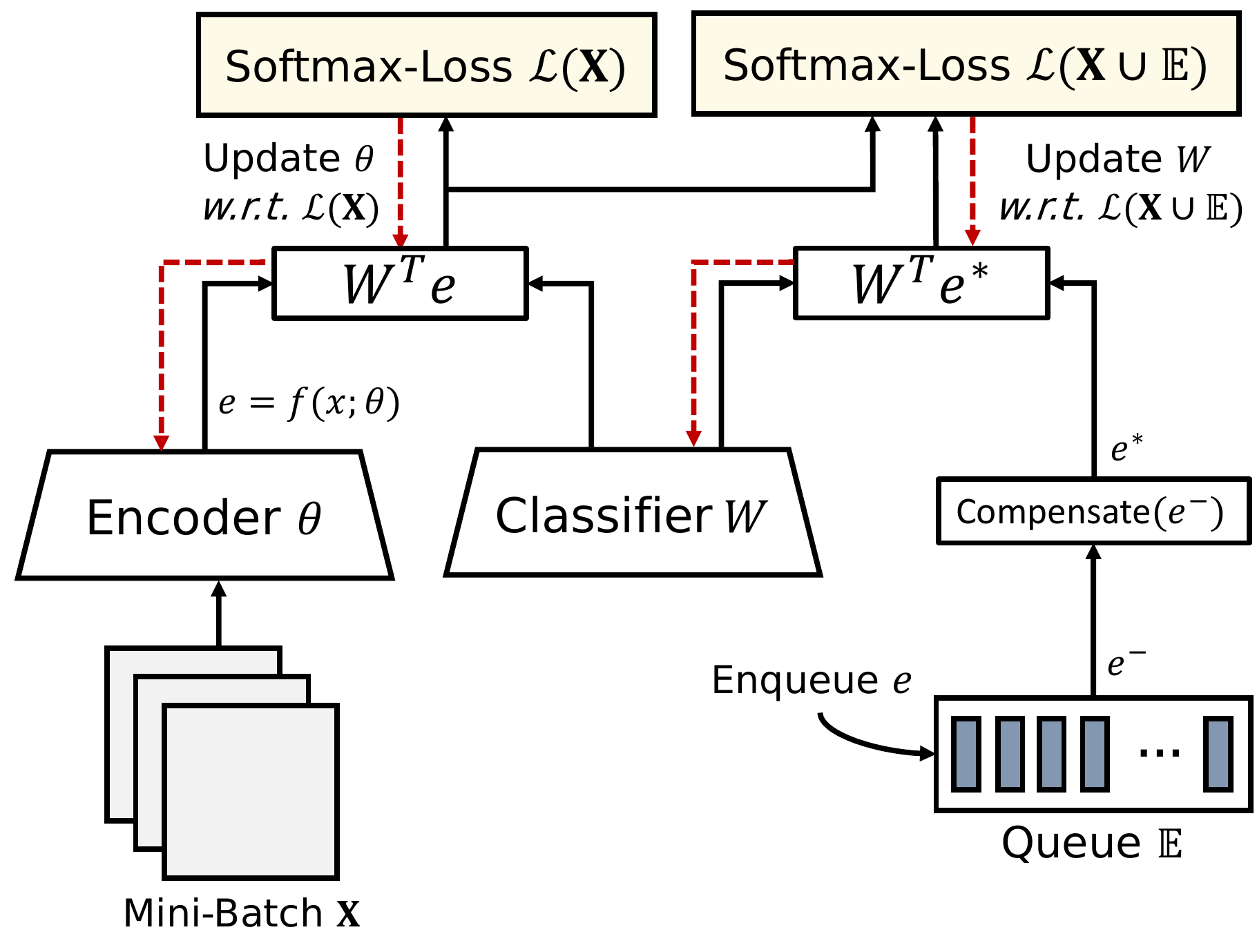}
		}	
	\end{center}
	\caption{
	(a) In typical mini-batch learning, the parameter $\theta$ of encoder $f$ and the parameter $W$ of linear classifier are optimized on a small mini-batch $\mathbf{X}$.
	(b) In the proposed method, the parameter of the encoder is optimized on a small mini-batch, but the parameter of the classifier is optimized on both the mini-batch and the large queue $\mathbb{E}$ that contains embedding vectors $e^{\text{-}}$ of past iterations.
	}
	\label{fig:teaser}
\end{figure}

Most previous studies focus on improving the discriminative power of an embedding space, because face recognition models are evaluated on independent datasets that include unseen identities.
The mainstream of recent studies \cite{ArcFace,SphereFace,NormFace,CosFace} is to introduce a new objective function to maximize inter-class discriminability and intra-class compactness; 
they try to consider all identities by referring an identity-representative vector, which is the weight vector of the last fully-connected layer for identity classification.

However, conventional methods still have difficulty in covering a massive set of identities at once, because these methods use a small mini-batch (Fig. \ref{fig:teaser_a}) much less than the number of identities due to memory constraints.
Inspecting tens of thousands of identities with the mini-batch of the small size requires numerous iterations, and this complicates the task of learning optimal decision boundaries in an embedding space while considering all of the identities, comprehensively.
Increasing the size of the mini-batch may alleviate some of the problem, but in general, this solution is impractical because of memory constraints; it also does not guarantee improved accuracy \cite{goyal2017accurate,hoffer2017train,keskar2016large,you2017scaling}.

We propose a novel method, called \textit{BroadFace}, which is a learning process to consider a massive set of identities, comprehensively (Fig. \ref{fig:teaser_b}).
BroadFace has a large queue to keep a massive number of embedding vectors accumulated over past iterations.
Our learning process increases the optimality of decision boundaries of the classifier by considering the embedding vectors of both a given mini-batch and the large queue for each iteration.
The parameters of the model are updated iteratively, so after a few iterations the error of enqueued embedding vectors gradually increases.
Therefore, we introduce a compensation method that reduces the expected error between the current and enqueued embedding vectors by referencing the difference of the identity-representative vectors of current and past iterations.
Our BroadFace has several advantages:
(1) the identity-representative vectors are updated with a large number of embedding vectors to increase the portion of the training set that is considered for each iteration,
(2) the optimality of the model is increased on the entire dataset by referring to the globally well-optimized identity-representative vectors,
(3) the learning process is accelerated.
We summarize the contributions as follows:
\begin{itemize}[leftmargin=+.2in,label=$\bullet$]
    \item
    We propose a new way that allows an embedding space to distinguish numerous identities in a broad perspective by learning identity-representative vectors from a massive number of instances.
    \item
    We perform extensive ablation studies on its behaviors, and experiments on various datasets to show the effectiveness of the proposed method, and to empirically prove the validity of our compensation method.
    \item
    BroadFace can be easily applied on many existing face recognition methods to obtain a significant improvement.
    Moreover, during inference time, it does not require any extra computational burden.
\end{itemize}

%% file: section/relatedworks.tex
\section{Related Works}

Recent studies of face recognition tend to introduce a new objective function that learns an embedding space by exploiting an identity-representative vector.
NormFace \cite{NormFace} reveal that optimization using cosine similarity between identity-representative vectors and embedding vectors is more effective than optimization using the inner product.
To increase the discriminative abilities of learned features, SphereFace \cite{SphereFace}, CosFace \cite{CosFace} and ArcFace \cite{ArcFace} adopted different kinds of margin into the embedding space.
Futhermore, some works adopted an additional loss function to regulate the identity-representative vectors.
RegularFace \cite{regularface} minimized a cosine similarity between identity-representative vectors, and UniformFace \cite{uniformface} equalized distances between all the cluster centers.
However, those methods can suffer from an enormous number of identities and instances because they are based on a mini-batch learning.
Our BroadFace overcomes the limitation of a mini-batch learning, and, it can be easily applied on those face recognition methods.

In terms of preserving knowledge of model on previously visited data, the continual learning \cite{hou2019learning,li2017learning} shares the similar concept with BroadFace.
However, BroadFace is different from continual learning, as BroadFace preserves knowledge of previous data from the same dataset while continual learning preserves knowledge of previous data from different datasets.

%% file: section/method.tex
\begin{figure}[t]
    \centering
    \includegraphics[width=12.3cm]{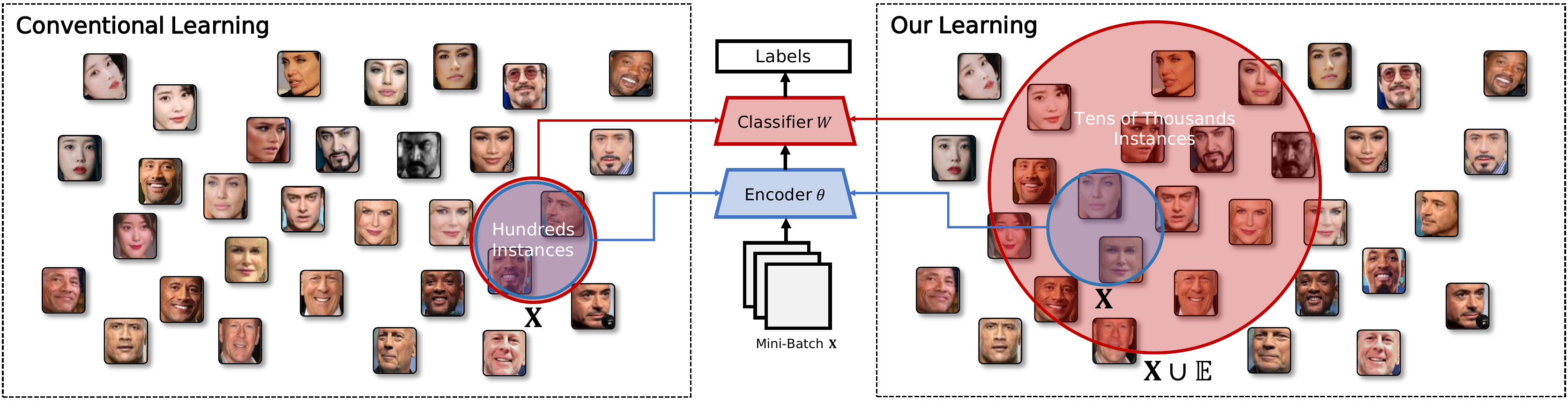}
	\caption{
Our learning refers more instances to learn the classifier in training stage.
	}  
	\label{fig:concept}
\end{figure}

\section{Proposed Method}
We describe the widely-adopted learning scheme in face recognition, and then illustrate the proposed BroadFace in detail. 

\subsection{Typical Learning}

\noindent\textbf{Learning of Face Recognition.}
In general, a face recognition network is divided into two parts:
(1) an encoder network that extracts an embedding vector from a given image
and (2) a linear classifier that maps an embedding vector into probabilities of identities.
An evaluation is performed by comparing embedding vectors on images of unseen identities, so the classifier is discarded at the inference stage.
Here, $f$ is the encoder network that extracts a $D$-dimensional embedding vector $e$ from a given image $x$: $e = f(x; \theta)$ with a model parameter $\theta$.
The linear classifier performs classification for $C$ identities from an embedding vector
$e$ with a weight matrix $W$ of $C$$\times$$D$-dimensions.
For a mini-batch $\mathbf{X}$, the objective function such as a variant of angular softmax losses \cite{ArcFace,SphereFace,NormFace} is used to optimize the encoder and the classifier: 
\begin{equation}
    \mathcal{L}(\mathbf{X}) = \frac{1}{|\mathbf{X}|}\sum_{i\in \mathbf{X}}{l(e_i, y_i)},
    \label{eq:general_form1}
\end{equation}
\begin{equation}
    l(e_i, y_i) = - \log \frac{\text{exp}(\hat{{W}}^{T}_{y_i}
    \hat{e}_i)}{\sum_{j=1}^{C}{\text{exp}(\hat{{W}}^{T}_{j}\hat{e}_j)}},
    \label{eq:general_form2}
\end{equation}
where ${y}_i$ is an labeled identity of ${x}_i$ and $\hat{\cdot}$ indicates that a given vector is $L_2$ normalized (\eg, $\norm{\hat{{e}}}_2=1$).
In Eq. \ref{eq:general_form2}, $\hat{W}_{y_i}$ acts as an representative instance of the given identity $y_i$ that maximizes the cosine similarity with an embedding vector $e_i$.
Thus, $\hat{W}_{y}$ can be regarded as the identity-representative vector, which is the expectation of instances belong to $y$: 
\begin{equation}
\hat{W}_y = E_{x}\left[\hat{e}_{i}\big|y_i = y\right].
\label{eq:expected}
\end{equation}

\noindent\textbf{Limitations of Mini-batch.}
The parameters of the model are updated in an iterative process that considers a mini-batch that contains only a small portion of the entire dataset for each step (Fig. \ref{fig:concept}).
However, use of a small mini-batch may not represent the entire distribution of training datasets.
Moreover, in face recognition, the number of identities is very large and each mini-batch only contains few of them; for example, MSCeleb-1M \cite{msceleb1m} has 10M images of 100k celebrities.
Therefore, each parameter update of a model can be biased on a small number of identities, and this restriction complicates the task of finding optimal decision boundaries.
Enlarging the mini-batch size may mitigate the problem, but this solution requires heavy computation of the encoder, proportional to the batch-size.

\subsection{BroadFace}

We introduce BroadFace, which is a simple yet effective way to cover a large number of instances and identities.
BroadFace learns globally well-optimized identity-representative vectors from a massive number of embedding vectors (Fig. \ref{fig:concept}).
For example, on a single Nvidia V100 GPU, the size of a mini-batch for ResNet-100 is at most 256, whereas BroadFace can utilize more than 8k instances at once.
The following describes each step.

\noindent\textbf{(1) Queuing Past Embedding Vectors.}
BroadFace has two queues of pre-defined size: $\mathbb{E}$ stores embedding vectors; $\mathbb{W}$ stores identity-representative vectors from the past.
For each iteration, after model update, embedding vectors of a given mini-batch $\{e_i\}_{i\in\mathbf{X}}$ are enqueued to $\mathbb{E}$, and corresponding identity-representative vectors $\{W_{y_i}\}_{i\in\mathbf{X}}$ of each instance are enqueued to $\mathbb{W}$.
By referring to the past embedding vectors in the queue to compute the loss $\mathcal{L}(\mathbf{X}\cup\mathbb{E})$, the network increase the number of instances and identities explored at each update. 

\begin{figure}[t]
	\begin{center}
		\subfloat[without Compensation]{
			\label{fig:compensate1}
			\includegraphics[width=5.3cm]{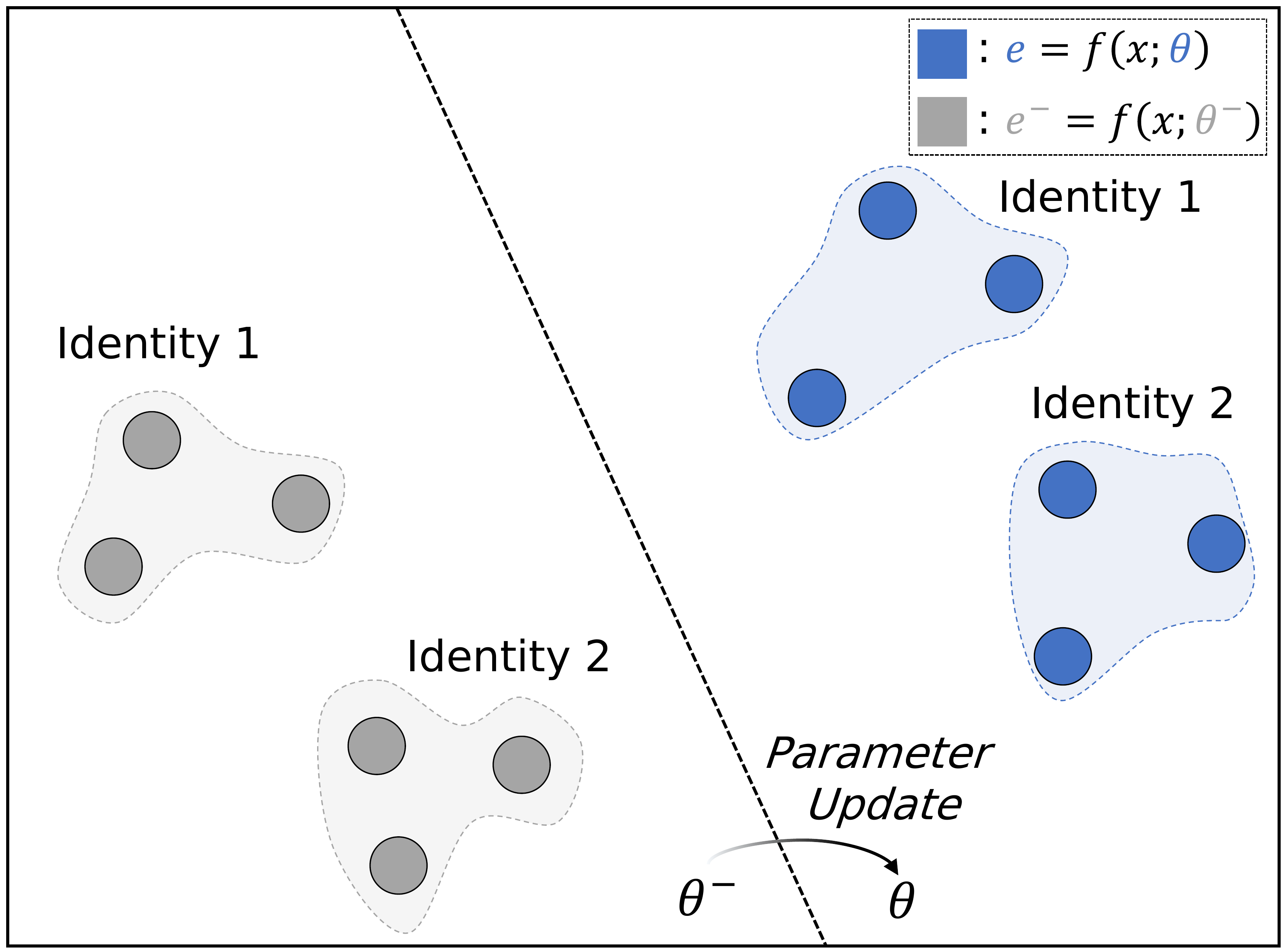}
		}
		\subfloat[with Compensation]{
			\label{fig:compensate2}
			\includegraphics[width=5.3cm]{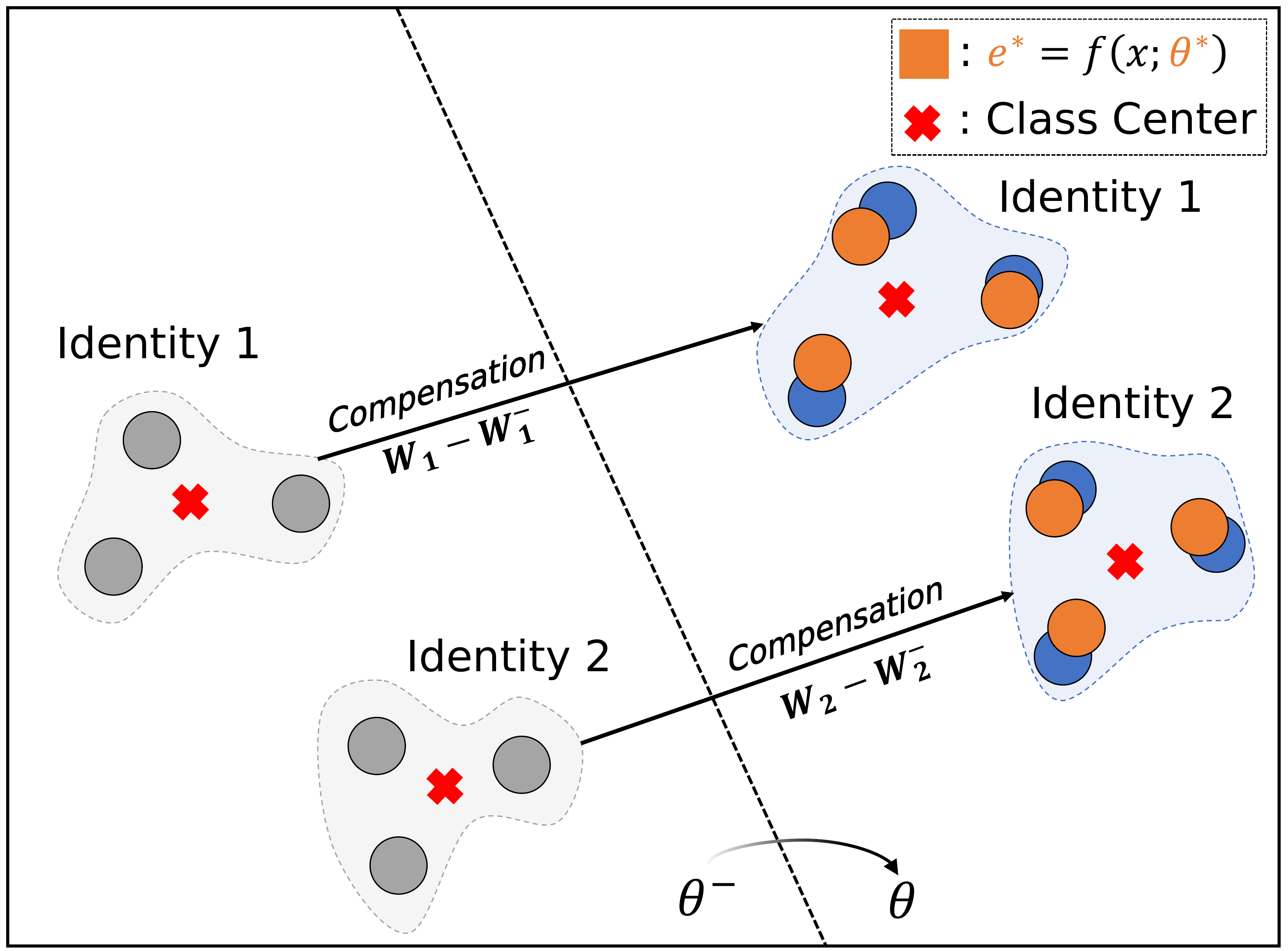}
		}	
	\end{center}
	\caption{
 	(a) enqueued embedding vectors (gray circles) at the past are further away from the embedding vectors (blue circles) at the current iteration due to the parameter update and this indicates significant errors.
 	(b) the compensated embedding vectors (orange circles) closely approach to the embedding vectors at the current iteration, by considering the difference between the identity-representative vectors (class centers) at the past and the current iteration.
	}
	\label{fig:compensate}
\end{figure}

\noindent\textbf{(2) Compensating Past Embedding Vectors. }
As the model parameter $\theta$ of the encoder is updated over iterations, past embedding vectors $e^{\text{-}}\in\mathbb{E}$ conflict with the embedding space of the current parameter (Fig. \ref{fig:compensate1}); $\epsilon = e - e^{\text{-}}$ where $\theta^{\text{-}}$ is the past parameter of the encoder and $e^{\text{-}}=f(x;\theta^{\text{-}})$.
The magnitude of the error $\epsilon$ is relatively small when few iterations have been passed from the past.
However, the error is gradually accumulated over iterations and the error hinders appropriate training.
We introduce a compensation function $\rho(y)$ for each identity to reduce the errors as an additive model; $e_{i}^{*} = e^{\text{-}}_i + \rho(y)$ where $e_{i}^{*}$ is a compensated past embedding vector to a current embedding vector (Fig. \ref{fig:compensate2}).
The compensation function should minimize an expected squared error $J$ between the current embedding vectors and the compensated past embedding vectors that belong to $y$: \begin{equation}
\begin{aligned}
    \text{minimize } J\big(\rho(y)\big) = & E_x \left[ \big(e_i^{*} - e_i\big)^2 \big | y_i = y \right], \\
    = & E_x \left[ \big(e_i^{\text{-}} + \rho(y) - e_i\big)^2 \big | y_i = y \right]. 
\end{aligned}
\end{equation}
The partial derivative of $J$ with respect to $\rho(y)$ is:
\begin{equation}
    \frac{\partial J}{\partial \rho(y)}=
    E_x \Big[ 2\big(e_i^{\text{-}} + \rho(y) - e_i\big) \big| y_i = y\Big]. 
\end{equation}
Thus, the optimal compensation function is the difference between the expectations of current embedding vectors and past embedding vectors:
\begin{equation}
\begin{aligned}
    \rho(y) & = E_x \left[ e_i \big | y_i = y \right] - E_x\left[e_i^{\text{-}} \big | y_i = y \right], \\
            & \approx \lambda (W_{y} - W^{\text{-}}_{y}),
\end{aligned}
\end{equation}
where $W^{\text{-}}_y\in\mathbb{W}$ is an identity-representative vector, which is enqueued to the queue during the same iteration as $e^{\text{-}}_{i}\in\mathbb{E}$.
As explained in Eq. \ref{eq:expected}, the identity-representative vector and the expected embedding vector point in the same direction when the vectors are projected onto a hyper-sphere, but the vectors are different in scale.
Thus, we deploy a simple normalization term per each instance to adjust these scales: $\lambda={\norm{e^{\text{-}}_i}}/\norm{W^{\text{-}}_{y_i}}$.
Then the compensated embedding vector $e^{*}$ is computed as: 
\begin{equation}
    e_{i}^{*} = e_{i}^{\text{-}} + \frac{\norm{e^{\text{-}}_i}}{\norm{W^{\text{-}}_{y_i}}} (W_{y_i} - W^{\text{-}}_{y_i}).
\end{equation}
In empirical studies, the compensation function reduces the error significantly.

\begin{figure}[t]
    \centering
    \includegraphics[width=12.3cm]{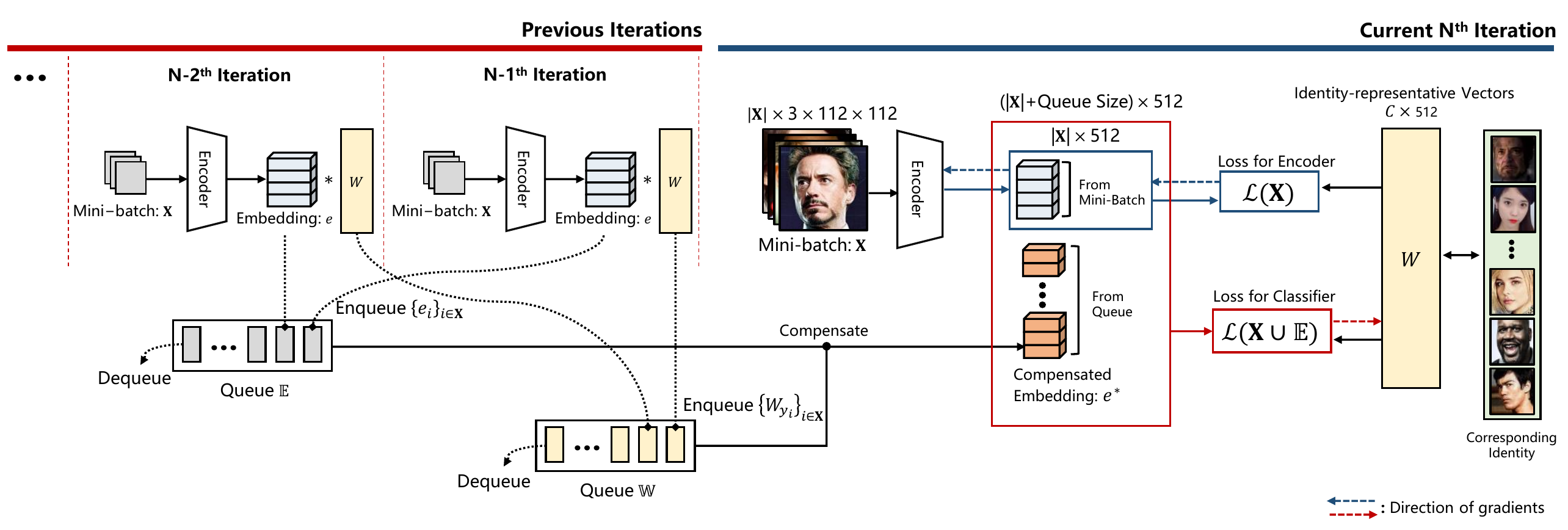}
	\caption{
    Learning process of the proposed method.
    BroadFace deploys large queues to store embedding vectors and their corresponding identity-representative vectors per iteration.
    The embedding vectors of the past instances stored in the queues are used to compute loss for identity-representative vectors. 
    BroadFace effectively learns from tens of thousands of instances for each iteration.
	}  
	\label{fig:network}
\end{figure}

\noindent\textbf{(3) Learning from Numerous Embedding Vectors.}
By executing the preceding two steps, BroadFace generates additional large-scale embedding vectors from the past.
In our method, the encoder is trained on a mini-batch as before and the classifier is trained on both a mini-batch and the additional embedding vectors.  
The objective functions for the encoder and the classifier are defined as:
\begin{equation}
    \mathcal{L}_{\text{encoder}}(\mathbf{X}) = \frac{1}{|\mathbf{X}|}
    \left \{
    \sum_{i\in \mathbf{X}}{l(e_i)}
    \right \},\\
    \label{eq:divide_objective1}
\end{equation}
\begin{equation}
     \mathcal{L}_{\text{classifier}}(\mathbf{X}\cup\mathbb{E}) =  \frac{1}{|\mathbf{X} \cup \mathbb{E}|}
    \left \{
    \sum_{i\in \mathbf{X}}{l(e_i)}
    +
    \sum_{j\in \mathbb{E}}{l(e^{*}_j )}
    \right \}.
    \label{eq:divide_objective2}
\end{equation}
The parameter $\theta$ of the encoder is updated \wrt $\mathcal{L}_{\text{encoder}}(\mathbf{X})$ while the parameter of the classifier $W$ is updated \wrt $\mathcal{L}_{\text{classifier}}(\mathbf{X}\cup\mathbb{E})$.
The large number of embedding vectors in the queue helps to learn highly precise identity-representative vectors that show reduced bias on a mini-batch and increased optimality on the entire dataset.
The precise identity-representative vectors can accelerate the learning procedure.
Moreover, our method can be easily implemented by adding several queues in the learning process (Fig. \ref{fig:network}) and significantly improves accuracy in face recognition without any computational cost at inference stage. 

\subsection{Discussion}
\begin{figure}[t]
	\begin{center}
		\subfloat[Averaged Error]{
			\label{fig:vis_com1}
			\includegraphics[width=5.5cm]{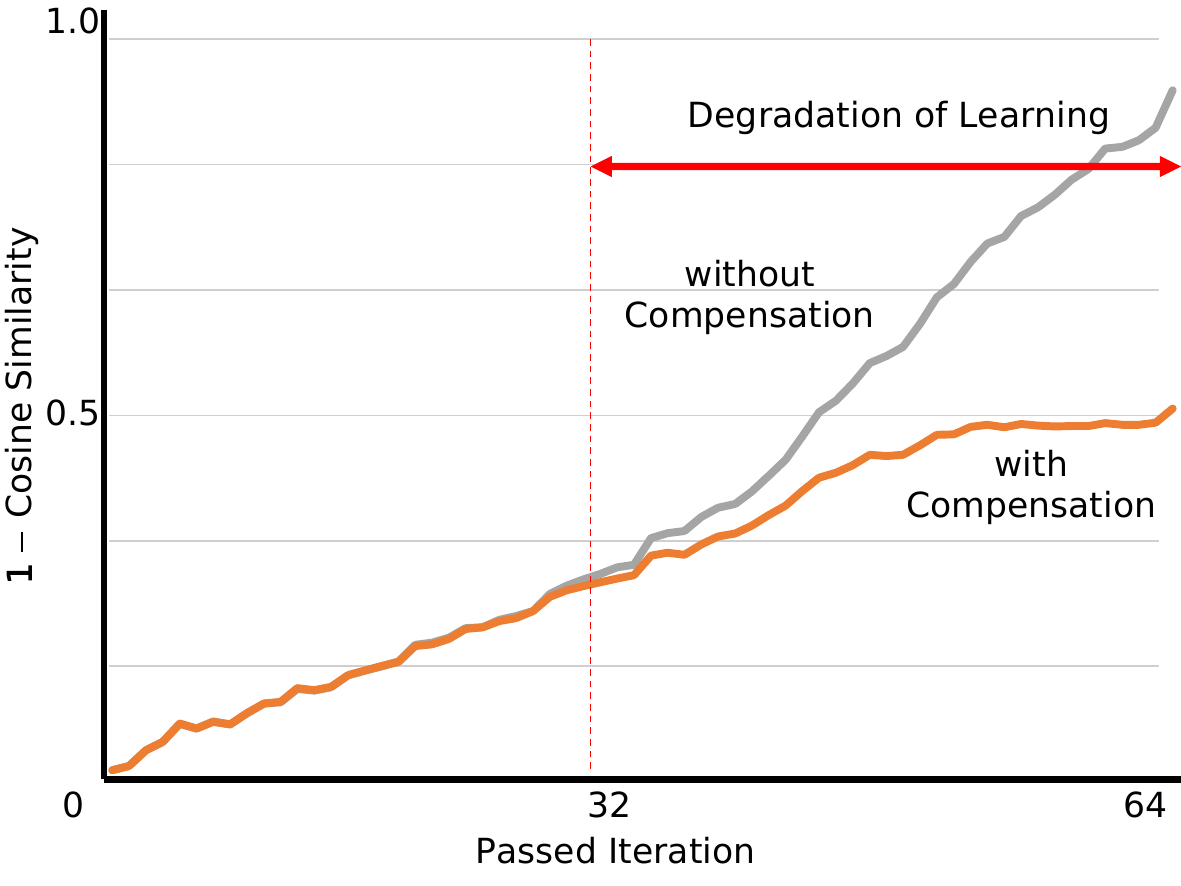}
		}
		\subfloat[Scatter Plot]{
			\label{fig:vis_com2}
			\includegraphics[width=5.5cm]{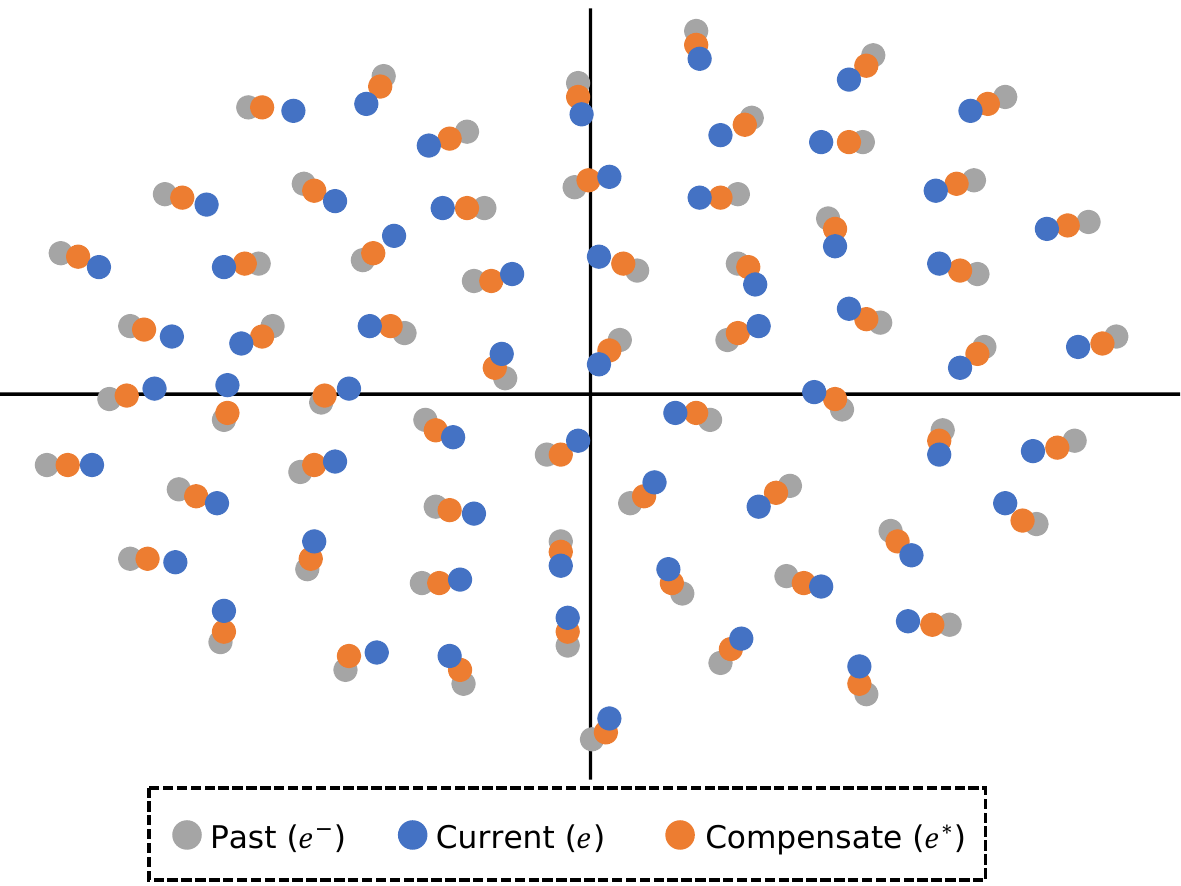}
		}	
	\end{center}
	\caption{
	(a) the average of the cosine errors between the embedding vectors at the current iteration and the past iteration with and without compensation.
	The errors are computed with randomly sampled instances over 64 iterations.
	(b) the scatter plot of the past (before 64 iterations), current and compensated embedding vectors for 64 instances.
	}
	\label{fig:vis_com}
\end{figure}

\noindent\textbf{Effectiveness of Compensation. }
We show that the compensation method is empirically effective.
After a small number of iterations, the error of the enqueued embedding vectors is also small and the compensation method is not necessary.
However, after a large number of iterations, the error increases and the compensation method becomes necessary to keep a large number of embedding vectors (Fig. \ref{fig:vis_com1}).
A large accumulated error may degrade the training process of the network (Fig. \ref{fig:img1} and Fig. \ref{fig:img2}).
We illustrate how the compensation function reduces the difference between past and current embedding vectors in 2-dimensional space by t-SNE \cite{t_SNE} (Fig. \ref{fig:vis_com2}).
The past embedding vectors approach to current embedding vectors after applying compensation.
This shows that the proposed compensation function works properly in practice.

\begin{figure}[t]
	\begin{center}
	    \includegraphics[width=5.5cm]{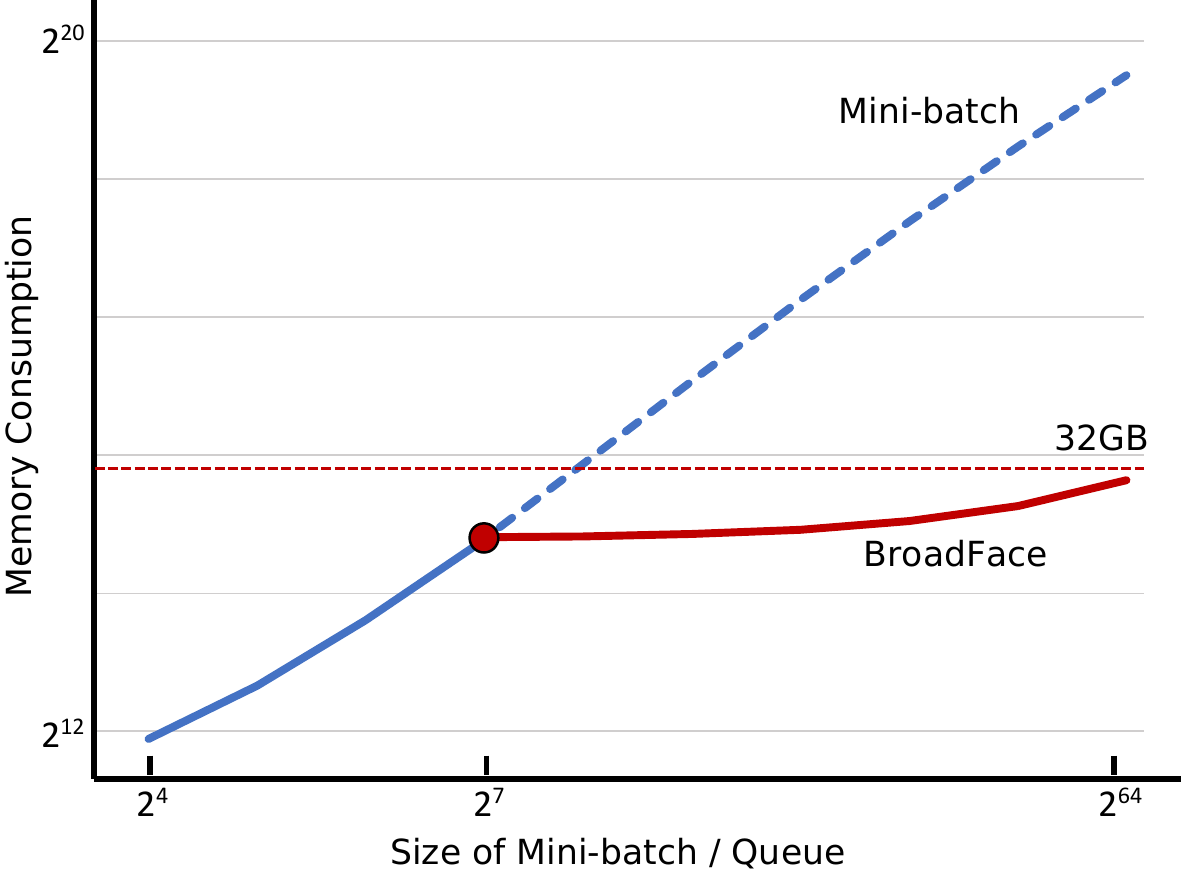}
	\end{center}
	\caption{
	Illustration on memory consumption of a conventional mini-batch learning (blue line) and the proposed BroadFace (red line) depending on the size of a mini-batch.
	The blue dotted line indicates memory consumption that is estimated for a large size of mini-batch by linear regression.
	}  
	\label{fig:memory}
\end{figure}

\noindent\textbf{Memory Efficiency.}
We compare BroadFace with enlarging the size of a mini-batch in terms of memory consumption (Fig. \ref{fig:memory}).
A na\"ive mini-batch learning requires a huge amount of memory to forward and backward the entire network.
The maximum size of a mini-batch is about $240$ instances when a model based on ResNet-100 is trained on NVidia V100 of 32 GB.
However, BroadFace requires only a matrix multiplication between the embedding vectors in $\mathbb{E}$ and the weight matrix $W$ (Eq.~\ref{eq:general_form2}).
The marginal computational cost of BroadFace enables a classifier to learn decision boundaries from a massive set of instances, \eg, 8192 instances for a single GPU.
Note that, enlarging the size of a mini-batch to 8192 requires about 952 GB of memory which is infeasibly large for a single GPU.

%% file: section/experiment.tex
\section{Experiments}

\subsection{Implementation Details}

\noindent\textbf{Experimental Setting.}
As pre-processing, we normalize a face image to $112 \times 112$ by warping a face-region using five facial points from two eyes, nose and two corners of mouth \cite{ArcFace,SphereFace,CosFace}.
A backbone network is ResNet-100 \cite{resnet} that is used in the recent works \cite{ArcFace,AFRN}.
After the \texttt{res5c} layer of ResNet-100, a block of batch normalization, fully-connected and batch normalization layers is deployed to compute a 512-dimensional embedding vector.
The computed embedding vectors and the weight vectors of the linear classifier are $L_2$-normalized and trained by the ArcFace~\cite{ArcFace}.
Our model is trained on $4$ synchronized NVidia V100 GPUs and a mini-batch of 128 images is assigned for each GPU.
The queue of BroadFace stores up to 8,192 embedding vectors accumulated over 64 iterations for each GPU, thus the total size of the queues is 32,768 for 4 GPUs. 
To avoid abrupt changes in the embedding space, the network of BroadFace is trained from the pre-trained network that is trained by the softmax based loss \cite{ArcFace}.
We adopted stochastic gradient descent (SGD) optimizer, and a learning rate is set to $5\cdot10^{-3}$ for the first 50k, $5\cdot10^{-4}$ for the 20k, and $5\cdot10^{-5}$ for the 10k with a weight decay of $5\cdot10^{-4}$ and a momentum of $0.9$.

\noindent\textbf{Datasets.}
All the models are trained on MSCeleb-1M \cite{msceleb1m}, which is composed of about 10M images for 100k identities.
We use the refined version \cite{ArcFace}, which contains 3.8M images for 85k identities by removing the noisy labels of MSCeleb-1M.
For the test, we perform evaluations on the following various datasets:
\begin{itemize}[label=$\bullet$]
    \item
    Labeled Faces in the Wild (LFW) \cite{LFW} contains 13k images of faces that are collected from web for 5,749 different individuals.
    Cross-Age LFW (CALFW) \cite{CALFW} provides pairs with age variation, and Cross-Pose (CPLFW) \cite{CPLFWTech} provides pairs with pose variation from the images of LFW.
    \item
    YouTube Faces (YTF) \cite{YTF} contains 3,425 videos of 1,595 different people.
    \item
    MegaFace \cite{kemelmacher2016megaface} contains more than 1M images from 690k identities to evaluate recognition-accuracy with enormous distractors.
    \item
    Celebrities in Frontal-Profile (CFP) \cite{CFP_FP}
    contains 500 subjects; each subject has 10 frontal and 4 profile images.
    \item
    AgeDB-30 \cite{age_db}, which contains 12,240 images of 440 identities with age variations, is suitable to evaluate the sensitivity of a given method in age variation.
    \item
    IARPA Janus Benchmark (IJB) \cite{IJB_C,IJB_B}, which is designed to evaluate unconstrained face recognition systems,
    is one of the most challenging datasets in public.
    IJB-B~\cite{IJB_B} is composed of 67k face images, 7k face videos and 10k non-face images.
    IJB-C~\cite{IJB_C}, which adds additional new subjects with increased occlusion and diversity of geographic origin to IJB-B, is composed of 138k face images, 11k face videos and 10k non-face images. 
\end{itemize}

\subsection{Evaluations on Face Recognition}
\label{sec:eval_fr}
We conduct experiments on the described various datasets to show the effectiveness of the proposed method.

\begingroup
\newcolumntype{C}[1]{>{\centering\let\newline\\\arraybackslash\hspace{0pt}}m{#1}}
 \begin{table}[th!]
    \footnotesize
    \begin{center}
     \caption{Verification accuracy ($\%$) on LFW and YTF.}
	 \label{tab:lfw}
     \end{center}
     \centering
	\def\arraystretch{1.1}
    \begin{tabular}{p{3cm}|C{1.3cm}|C{1.3cm}||p{3cm}|C{1.3cm}|C{1.3cm}}
    \hline
    Method & LFW & YTF & Method & LFW & YTF \\
    \hline
    \hline
    DeepID \cite{contrastive_loss}      & 99.47 & 93.2 &
    DeepFace \cite{deepface}            & 97.35 & 91.4 \\
    VGGFace \cite{vggface}              & 98.95 & 97.3 &
    FaceNet \cite{FaceNet}              & 99.64 & 95.1 \\
    CenterLoss \cite{centerloss}        & 99.28 & 94.9 &
    RangeLoss \cite{range_loss}         & 99.52 & 93.7 \\
    MarginalLoss \cite{marginal_loss}   & 99.48 & 95.9 &
    SphereFace \cite{SphereFace}        & 99.42 & 95.0 \\
    RegularFace \cite{regularface}      & 99.61 & 96.7 &
    CosFace \cite{CosFace}              & 99.81 & 97.6 \\
    UniformFace \cite{uniformface}      & 99.80 & 97.7 &
    AFRN \cite{AFRN}                    & 99.85 & 97.1 \\
    ArcFace \cite{ArcFace}              & 99.83 & 97.7 &
    BroadFace                           & $\mathbf{99.85}$ & $\mathbf{98.0}$ \\
    \hline
    \end{tabular}
\end{table}
\endgroup
\begingroup
\newcolumntype{C}[1]{>{\centering\let\newline\\\arraybackslash\hspace{0pt}}m{#1}}
\begin{table}[th!]
    \footnotesize
    \begin{center}
     \caption{Verification accuracy ($\%$) on CALFW, CPLFW, CFP-FP and AgeDB-30.}
	 \label{tab:agedb}
	 \end{center}
	 \centering
	 \small
	\def\arraystretch{1.1}
    \begin{tabular}{p{3cm}|C{1.5cm}|C{1.5cm}|C{1.5cm}|C{1.5cm}}
    \hline
    {Method} & {CALFW} & {CPLFW} & {CFP-FP} & {AgeDB-30} \\
    \hline
    \hline
    CenterLoss \cite{centerloss}    & 85.48 & 77.48 & - & -\\
    SphereFace \cite{SphereFace}    & 90.30 & 81.40 & - & -\\
    VGGFace2 \cite{vggface2}        & 90.57 & 84.00 & - & -\\
    CosFace \cite{CosFace}          & 95.76 & 92.28 & 98.12 & 98.11\\
    ArcFace \cite{ArcFace}          & 95.45 & 92.08 & 98.27 & 98.28 \\
    BroadFace                       & $\mathbf{96.20}$ & $\mathbf{93.17}$ & $\mathbf{98.63}$ & $\mathbf{98.38}$\\
    \hline
    \end{tabular}
 \end{table}
 \endgroup

\begingroup
\newcolumntype{C}[1]{>{\centering\let\newline\\\arraybackslash\hspace{0pt}}m{#1}}
\begin{table}[th!]
    \footnotesize
    \begin{center}
     \caption{Identification and verification evaluation on MegaFace \cite{kemelmacher2016megaface}.
     Ident indicates rank-1 identification accuracy ($\%$) and Verif indicates a true accept rate ($\%$) at a false accept rate of 1e-6.
     }
	 \label{tab:mega}
	 \end{center}
	 \centering
	\def\arraystretch{1.1}
    \begin{tabular}{p{3cm}|C{1.6cm}|C{1.6cm}|C{1.6cm}|C{1.6cm}}
    \hline
    \multirow{2}{*}{Method} &  \multicolumn{2}{c|}{MF-Large} &  \multicolumn{2}{c}{MF-Large-Refined \cite{ArcFace}}  \\
    \cline{2-5}
    & {Ident} & {Verif} & {Ident} & {Verif} \\
    \hline
    \hline
    RegularFace \cite{regularface}      & 75.61 & 91.13 & - & - \\
    UniformFace \cite{uniformface}      & 79.98 & 95.36 & - & - \\
    SphereFace \cite{SphereFace}        & - & - & 97.91 & 97.91 \\
    AdaptiveFace \cite{adaptiveface}    & - & - & 95.02 & 95.61 \\
    CosFace \cite{CosFace}              & 80.56 & 96.56 & 97.91 & 97.91 \\
    ArcFace \cite{ArcFace}              & 81.03 & 96.98 & 98.35 & 98.49 \\
    BroadFace                           & $\mathbf{81.33}$ & $\mathbf{97.56}$ & $\mathbf{98.70}$ & $\mathbf{98.95}$ \\
    \hline
    \end{tabular}
\end{table}
\endgroup

 \begingroup
\newcolumntype{C}[1]{>{\centering\let\newline\\\arraybackslash\hspace{0pt}}m{#1}}
\begin{table}[th!]
    \footnotesize
   \begin{center}
    \caption{Verification evaluation with a True Accept Rate at a certain False Accept Rate (TAR@FAR) from 1e-4 to 1e-6 on IJB-B and IJB-C. 
    $^{\dag}$ denotes BroadFace trained by CosFace \cite{CosFace}.}
    \label{tab:ijb}
    \end{center}
    \centering
	\def\arraystretch{1.1}
   \begin{tabular}{p{3cm}|c|c|c|c|c|c}
   \hline
   \multirow{2}{*}{Method} &  \multicolumn{3}{c|}{IJB-B} &  \multicolumn{3}{c}{IJB-C}  \\
   \cline{2-4}\cline{5-7}
    & \scalebox{0.9}{FAR=1e-6} & \scalebox{0.9}{FAR=1e-5} & \scalebox{0.9}{FAR=1e-4} & \scalebox{0.9}{FAR=1e-6} & \scalebox{0.9}{FAR=1e-5} & \scalebox{0.9}{FAR=1e-4}  \\
    \hline\hline
    VGGFace2 \cite{vggface2}            & - & 0.671 & 0.800 &  - & 0.747 & 0.840 \\
    CenterFace \cite{centerloss}        & - & - & - & - & 0.781 & 0.853 \\
    ComparatorNet \cite{ComparatorNet}  & - & -     & 0.849 &  - & - & 0.885 \\
    PRN \cite{PRN}                      & - & 0.721 & 0.845 &  - & - & - \\
    AFRN \cite{AFRN}                    & - & 0.771 & 0.885 &  - & 0.884 & 0.931 \\
    \hline
    CosFace \cite{CosFace}              & 0.3649 & 0.8811 & 0.9480 & 0.8591 & 0.9410 & 0.9637 \\
    BroadFace$^\dag$                 & 0.4092 & 0.8997 & $\mathbf{0.9497}$ & 0.8596 & $\mathbf{0.9459}$ & $\mathbf{0.9638}$ \\  
    \hline
    ArcFace \cite{ArcFace}              & 0.3828 & 0.8933 & 0.9425 & 0.8906 & 0.9394 & 0.9603 \\
    BroadFace                 & $\mathbf{0.4653}$ & $\mathbf{0.9081}$ & 0.9461 & $\mathbf{0.9041}$ & 0.9411 & 0.9603 \\
   \hline
   \end{tabular}
\end{table}
\endgroup

\noindent\textbf{LFW and YTF}
are widely used to evaluate verification performance under the unrestricted environments.
LFW, which contains pairs of images, evaluates a model by comparing two embedding vectors of a given pair.
YTF contains videos that are sets of images; from the shortest clip of 48 frames to the longest clip of 6,070 frames. 
To compare a pair of videos, YTF compares a pair of video-representative embedding vectors that are averaged embedding vectors of images collected from each video.
Even though both datasets are highly-saturated in accuracy, our BroadFace outperforms other recent methods (Table \ref{tab:lfw}).

\noindent\textbf{CALFW, CPLFW, CFP-FP and AgeDB-30}
are also widely used to verify that methods are robust to pose and age variation.
CALFW and AgeDB-30 have multiple instances for same identity of different ages
and CPLFW and CFP-FP have multiple instances for same identity of different poses (frontal and profile faces).
BroadFace shows better verification-accuracy on all datasets (Table \ref{tab:agedb}).

\noindent\textbf{MegaFace}
is designed to evaluate both face identification and verification tasks under difficulty caused by a huge number of distractors.
We evaluate our BroadFace on Megaface Challenge 1 where the training dataset is more than 0.5 million images.
BroadFace outperforms the other top-ranked face recognition models for both face identification and verification tasks (Table \ref{tab:mega}).
On the refined MegaFace \cite{ArcFace}, where noisy labels are removed, BroadFace also surpasses the other models.

\noindent\textbf{IJB-B and IJB-C}
are the most challenging datasets to evaluate unconstrained face recognition.
We report BroadFace with CosFace \cite{CosFace} and BroadFace with ArcFace \cite{ArcFace} in verification task without any augmentations such as horizontal flipping in test time.
Our BroadFace shows significant improvements on all FAR criteria (Table \ref{tab:ijb}).
In IJB-B \cite{IJB_B}, BroadFace improves 8.25 percentage points on FAR=1e-6, 1.48 percentage points on FAR=1e-5 and 0.36 percentage points on FAR=1e-4 comparing to the results of ArcFace \cite{ArcFace}.

\subsection{Evaluations on Image Retrieval.}
\label{sec:eval_ir}

Both face recognition and image retrieval have the same goal that learn an optimal embedding space to compare a given pair of items such as face, clothes or industrial products.
To show that BroadFace is widely applicable on other applications, we compare BroadFace with recently proposed metric learning methods for image retrieval.

\noindent\textbf{Experimental Settings.} 
We use ResNet-50 \cite{resnet} that is pre-trained on ILSVRC 2012-CLS~\cite{ILSVRC15} as a backbone network.
We use ArcFace \cite{ArcFace} as a baseline objective function and set the size of the queue to 32k for BroadFace.
We follow the standard input augmentation and evaluation protocol \cite{song2016deep}.
We evaluate on two large datasets with a large number of classes similar to face-recognition:
In-Shop Clothes Retrieval (In-Shop) \cite{liuLQWTcvpr16DeepFashion} and Stanford Online Products (SOP) \cite{song2016deep}.

\begingroup
\newcolumntype{C}[1]{>{\centering\let\newline\\\arraybackslash\hspace{0pt}}m{#1}}
\begin{table}[t]
    \footnotesize
    \begin{center}
    \caption{Recall@K comparison with state-of-the-art methods. For fair comparison, we divide methods according to the dimension (Dim.) of an embedding vector. 
    The numbers under the datasets refer to recall at K.
    }
    \label{tab:image_retrieval}
    \end{center}
    \centering
	\def\arraystretch{1.1}
	\begin{tabular}{p{3.5cm}|C{1cm}|C{0.8cm}|C{0.8cm}|C{0.8cm}|C{0.8cm}|C{0.8cm}|C{0.8cm}|C{0.8cm}|C{0.8cm}}
   \hline
    \multirow{2}{*}{Methods} &  \multirow{2}{*}{Dim.} & \multicolumn{4}{c|}{In-Shop} & \multicolumn{4}{c}{SOP} \\
   \cline{3-6}\cline{7-10}
       & & 1 & 10 & 20 & 30 & 1 & 10 & $10^2$ & $10^3$  \\
    \hline\hline
    Margin~\cite{wu2017sampling} & 128 & - & - & - & - & 72.7 & 86.2 & 93.8 & 98.0\\
      MIC+Margin~\cite{Roth_2019_ICCV} & 128 & 88.2 & 97.0 & - & - & 77.2 & 89.4 & 95.6 & - \\
      DC~\cite{sanakoyeu2019divide} & 128 & 85.7 & 95.5 & 96.9 & 97.5 & 75.9 & 88.4 & 94.9 & 98.1  \\ 
      ArcFace \cite{ArcFace} & 128 & 84.1 & 94.9 & 96.2 & 96.9 & 73.3 & 86.4 & 93.2 & 97.1 \\ 
      BroadFace & 128 & $\mathbf{89.8}$ & $\mathbf{97.4}$ & $\mathbf{98.1}$ & $\mathbf{98.4}$ & $\mathbf{79.7}$ & $\mathbf{90.7}$ & $\mathbf{95.7}$ & $\mathbf{98.4}$ \\ 
      \cline{1-10}
      TML~\cite{Yu_2019_ICCV} & 512 & - & - & - & - & 78.0 & $\mathbf{91.2}$ & $\mathbf{96.7}$ & $\mathbf{99.0}$ \\
      NSM~\cite{zhai2018making} & 512 & 88.6 & $\mathbf{97.5}$ & $\mathbf{98.4}$ & $\mathbf{98.8}$ & 78.2 & 90.6 & 96.2 & -\\
      ArcFace \cite{ArcFace}  & 512 & 87.3 & 96.3 & 97.3 & 97.9 & 76.9 & 89.1 & 95.0 & 98.2 \\ 
      BroadFace & 512 & $\mathbf{90.1}$ & 97.4 & 98.1 & 98.4 & $\mathbf{80.2}$ & 91.0 & 95.9 & 98.4 \\ 
   \hline
   \end{tabular}
\end{table}
\endgroup

\noindent\textbf{In-Shop and SOP}
are the standard datasets in image retrieval.
In-Shop contains 11,735 classes of clothes.
For training, the first 3,997 classes with 25,882 images are used, and the remaining 7,970 classes with 26,830 images are split into query set gallery set for evaluation.
SOP contains 22,634 classes of industrial products.
For training, the first 11,318 classes with 59,551 images are used and the remaining 11,316 classes with 60,499 images are used for evaluation.
Our baseline models that are trained with ArcFace \cite{ArcFace} underperform comparing to the other state-of-the-art methods.
BroadFace significantly improves the recall of the baseline models and the improved model even outperforms the other methods (Table \ref{tab:image_retrieval}).

\subsection{Analysis of BroadFace}

\begingroup
\newcolumntype{C}[1]{>{\centering\let\newline\\\arraybackslash\hspace{0pt}}m{#1}}
\begin{table}[t]
    \footnotesize
	\begin{center}
    	\caption{
            Effects of BroadFace varying the size of the queue and the type of the backbone network on IJB-B dataset in face recognition.
            \\
    	}
    	\label{tab:abl}
		\centering
		\subfloat[Total Size of Queue\label{tab:abl_a} ]
		{
			\centering
			\def\arraystretch{1.1}
            \begin{tabular}[t]{p{3.6cm}|C{1.9cm}|C{1.9cm}|C{1.9cm}|C{1.9cm}}
			\hline
			 \multirow{3}{*}{Size of Queue}  &   \multicolumn{4}{c}{TAR} \\
			    \cline {2-5}
				&   \multicolumn{2}{c|}{FAR=1e-6} & \multicolumn{2}{c}{FAR=1e-5} \\
			    \cline {2-5}
				& \scalebox{0.8}{\parbox[][0.8cm][c]{1.9cm}{\centering without \newline Compensation}} & \scalebox{0.8}{\parbox[][0.8cm][c]{1.9cm}{\centering with \newline Compensation}} & \scalebox{0.8}{\parbox[][0.8cm][c]{1.9cm}{\centering without \newline Compensation}} & \scalebox{0.8}{\parbox[][0.8cm][c]{1.9cm}{\centering with \newline Compensation}} \\ 
				\hline
				\hline
				{0 (Baseline)}	& 0.3828 & 0.3828 & 0.8933 & 0.8933 \\ 
				{2048 ($512 \times 4$ GPUs) }	    	& 0.4310 & 0.4255  & 0.9061 & 0.9077 \\ 
				{8192 ($2048 \times 4$ GPUs) } 		& 0.4346 & 0.4394 & 0.9071 & 0.9085 \\ 
				{32768 ($8192\times4$ GPUs) } & 0.4259 & 0.4653 & 0.9078 & 0.9081 \\ 
				\hline
			\end{tabular} 
		}
		\\
		\centering
		\subfloat[Backbone Network\label{tab:abl_b}]
		{
			\centering
			\def\arraystretch{1.1}
			\begin{tabular}[t]{p{2.8cm}|C{1.0cm}|C{1.5cm}|C{1.5cm}|C{1.5cm}|C{1.5cm}|C{1.2cm}}
            \hline
				& \multirow{3}{*}{Dim.} &   \multicolumn{4}{c|}{TAR} & \multirow{3}{*}{GFlops}   \\
			    \cline{3-6}
				&  &   \multicolumn{2}{c|}{FAR=1e-6} & \multicolumn{2}{c|}{FAR=1e-5} &   \\
				 \cline{3-6}
				 &  &   \scalebox{0.9}{ArcFace} & \scalebox{0.9}{BroadFace} & \scalebox{0.9}{ArcFace} & \scalebox{0.9}{BroadFace} &   \\
				\hline\hline
				{MobileFaceNet \cite{MobileFaceNet}} & 128 & 0.3552 & 0.3665 & 0.8456 & 0.8458  & 0.9G \\
				{ResNet-18 \cite{resnet}}	& 128 & 0.3678 & 0.3808 & 0.8588 & 0.8638 & 5.2G \\
				{ResNet-34 \cite{resnet}}	& 512 & 0.3981 & 0.4325 & 0.8798 & 0.8828 & 8.9G \\
				ResNet-100 \cite{resnet}	& 512 & 0.3828 & 0.4653 & 0.8933 & 0.9081 & 24.1G \\
                \hline
			\end{tabular}
		}  
		\end{center}
\end{table}
\endgroup

\begin{figure}[t]
	\begin{center}
		\subfloat[Size of Queue]{
			\label{fig:img1}
			\includegraphics[width=5.5cm]{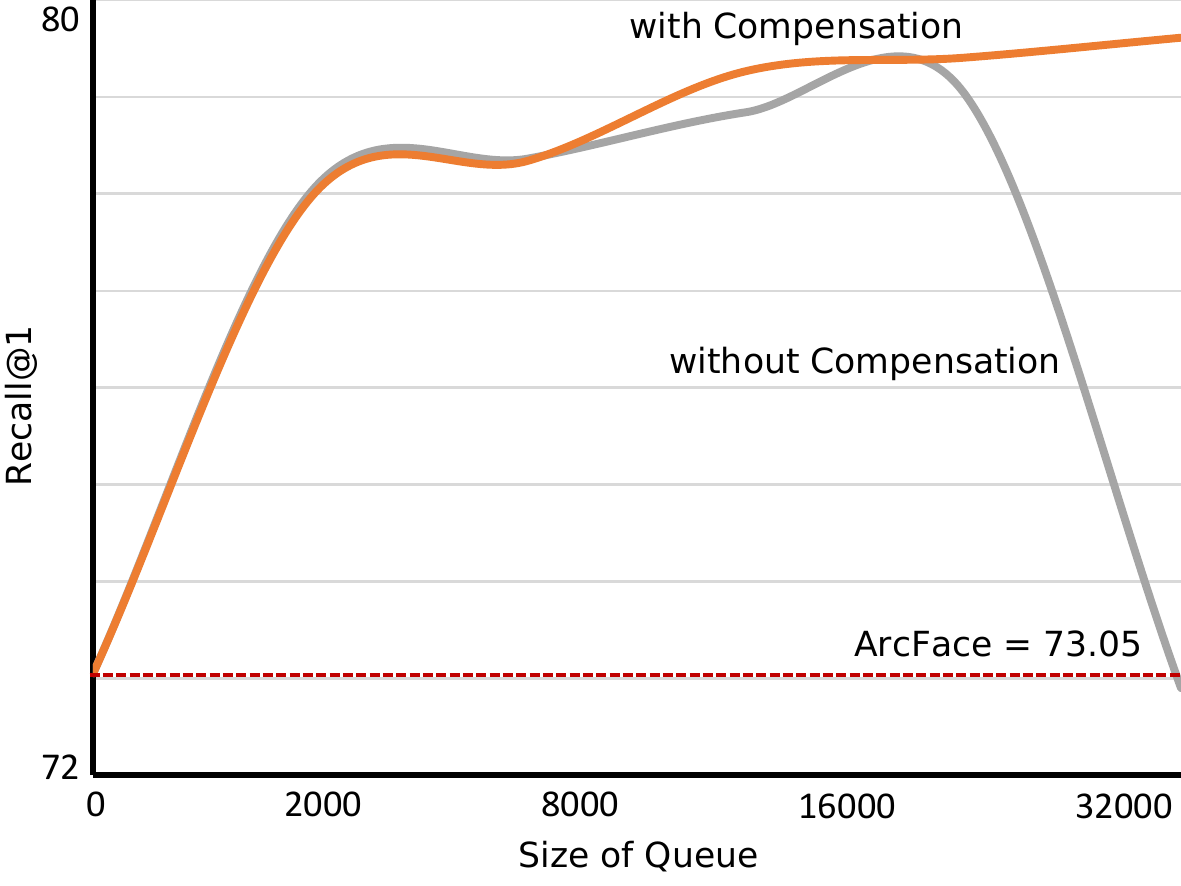}
		}
		\subfloat[Learning Acceleration]{
			\label{fig:img2}
			\includegraphics[width=5.5cm]{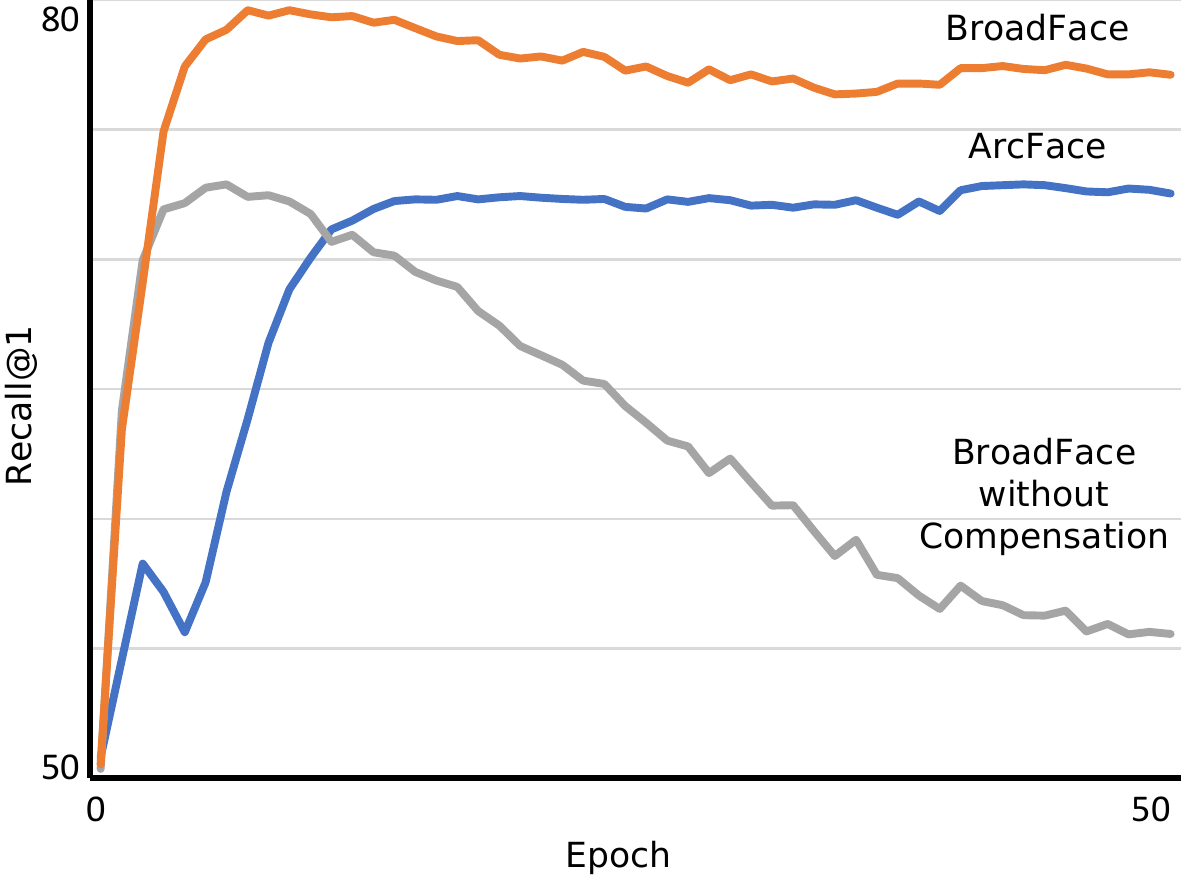}
		}	
	\end{center}
	\caption{
	(a) the recall depending on the size of the queue in BroadFace with and without our compensation function; the red line indicates the recall of ArcFace (baseline) on the test set.
(b) the learning curve for the test set when the size of the queue is 32k; ArcFace reaches the highest recall at the $45^{th}$ epoch, our BroadFace reaches the highest recall at the $10^{th}$ epoch, and the learning process collapses without our compensation function.
	}
	\label{fig:img}
\end{figure}

\noindent\textbf{Size of Queue.}
BroadFace has only one hyper-parameter, the size of the queue, to determine the maximum number of embedding vectors accumulated over past iterations.
Using the single parameter makes our method easy to tune and the parameter plays a very important role in determining recognition-accuracy.
As the size of the queue grows, the performance increases steadily (Table \ref{tab:abl_a}).
Especially, without our compensation method, accuracy degradation occurs when the size of the queue is significantly large.
However, our compensation method alleviates the degradation by correcting the enqueued embedding vectors.
We show another experiment on the enormous size of the queue from $0$ to $32,000$ in image retrieval (Fig. \ref{fig:img1}).
With the proposed compensation, the recall is consistently improved as the size of the queue is increased. 
However, without the proposed compensation, the recall of the models degrades when the size of the queue is more than 16k.

\noindent\textbf{Generalization Ability.}
Our BroadFace is generally applicable to any objective functions and any backbone networks.
We apply BroadFace to two widely used objective functions of CosFace \cite{CosFace} and ArcFace \cite{ArcFace}.
For both CosFace and ArcFace, BroadFace increases recognition-accuracy (Table \ref{tab:ijb}).
We also apply BroadFace to several backbone networks such as MobileFaceNet \cite{MobileFaceNet}, ResNet-18 \cite{resnet} and ResNet-34 \cite{resnet}.
We set the dimensions of embedding vector to 128 for light backbone networks such as MobileFaceNet and ResNet-18, and 512 for heavy backbone networks such as ResNet-34 and ResNet-100.
BroadFace is significantly effective for all backbone networks (Table \ref{tab:abl_b}).
In particular, ResNet-34 trained with BroadFace achieves comparable performance to ResNet-100 trained only with ArcFace, even though ResNet-34 has much less GFlops.

\noindent\textbf{Learning Acceleration.}
Our BroadFace accelerates the learning process of both face recognition and image retrieval.
In face recognition, many iterations are still needed to overcome a small gap of performance among the methods on the highly-saturated datasets.
Thus, we experiment the acceleration of the learning process in image retrieval to clearly show the effectiveness (Fig. \ref{fig:img2}).
Our BroadFace reaches peak performance much faster and higher than the baseline model.
Without our compensation method, the model gradually collapses.

%% file: section/conclusion.tex
\section{Conclusion}

We introduce a new way called BroadFace that allows an embedding space to distinguish numerous identities in a broad perspective by increasing the optimality of constructed identity-representative vectors.
BroadFace is significantly effective for face recognition and image retrieval where their datasets consist of numerous identities and instances.
BroadFace can be easily applied on many existing face recognition methods to obtain a significant improvement without any extra computational cost in the inference stage.